# THE TURING DECEPTION

David Noever[1] and Matt Ciolino[2]

PeopleTec, Inc., Huntsville, Alabama, USA
[1]david.noever@peopletec.com    [2]matt.ciolino@peopletec.com

## ABSTRACT

*This research revisits the classic Turing test and compares recent large language models such as ChatGPT for their abilities to reproduce human-level comprehension and compelling text generation. Two task challenges- summary and question answering- prompt ChatGPT to produce original content (98-99%) from a single text entry and sequential questions initially posed by Turing in 1950. We score the original and generated content against the OpenAI GPT-2 Output Detector from 2019, and establish multiple cases where the generated content proves original and undetectable (98%). The question of a machine fooling a human judge recedes in this work relative to the question of "how would one prove it?" The original contribution of the work presents a metric and simple grammatical set for understanding the writing mechanics of chatbots in evaluating their readability and statistical clarity, engagement, delivery, overall quality, and plagiarism risks. While Turing's original prose scores at least 14% below the machine-generated output, whether an algorithm displays hints of Turing's true initial thoughts (the "Lovelace 2.0" test) remains unanswerable.*

## KEYWORDS

*Large language Models (LM), Generative Pre-trained Transformer (GPT-3), Turing test*

The rise of generative text has brought the broader AI community to seek more challenging tests than the first landmark Turing test [1]. In short, Turing addressed two questions: "can a machine think?" and "how could one prove it?" The core method coined by Turing was his version of a question-and-answer format called the Imitation Game. Many refinements have appeared since 1950, including a notable 2014 test called the "Lovelace 2.0 Test" [2]. The Lovelace 2.0 test similarly lays out the requirements to conclude that a machine can create representative examples in art, literature, or any equally imaginative leap. As a corollary competition, the Loebner Prize awards the "most human computer" [3-4].

Critics of these tests point out that the AI community focuses on more exciting tasks than fooling human judges, particularly when the machine models can score themselves against beneficial benchmark goals such as summarizing complex topics in medicine, law, and government or assisting programmers with solving a computer coding problem [5]. Fundamentally, the Turing test plays on one of AI's most troubling goals: how to deceive unwitting humans? One contention of the present work is that as AI gets more sophisticated, the idea of deception diminishes somewhat, and the follow-on question of "how could one prove it?" becomes more important to examine [6-7]. One could argue that the modern Turing test proves more insightful in its call for establishing a language model in a scientific setting than whether it could fool an unknowing human [2, 8-9].

Figure 1 illustrates an actual question posed by Turing's 1950 paper [1], a challenging arithmetic question that, until the past five years, would have tripped up the best language models at the time, like GPT-2 [10-12]. The ironic outcome, however, is that Turing's paper (the human version, [1]) answers with the following incorrect answer [13]: *A: (Pause about 30 seconds and then answer) 105621.*

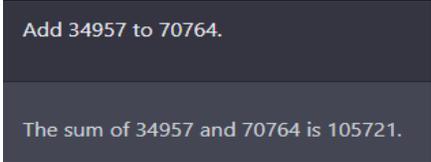

Figure 1. Question and answer sequence with ChatGPT, where the answer is correct and the question is from Turing's 1950 paper, which offered the incorrect (human) response without qualification.

Even the details added to the competitive efforts to pass the Turing test provide a successful exit when a five-minute keyboard conversation convinces a judging panel that a computer is a human more than 30% of the time. The University of Reading organized a 2014 event that reported fooling 33% of the panel, which they deemed as the first complete passing grade by a chatbot called Eugene Goostman, which simulated the persona of a 13-year-old Ukrainian boy [14]. Detractors of the event noted the pre-defined questions and topics, along with its short format of keyboard strokes only. In 2018, Google CEO Sundar Pichai provided a video of their latest computer assistant, Duplex, successfully booking a hair salon appointment as part of humans interacting unwittingly with a convincing machine [15]. While formally passing the Turing test might take several formats, The Big Think [15] concluded in 2022, "To date, no computer has decidedly passed the Turing AI test." Other researchers [5] reiterate the question as to whether deception seems worth all this trouble, particularly given the present wealth of valuable things that large language models can accomplish: "Aeronautical engineering texts do not define the goal of their field as 'making machines that fly so exactly like pigeons that they can fool other pigeons.'"

> ***Recognition of machine-generated narratives***
> *A complaint about machine-generated text generally stems from known model flaws, such as a tendency to lose context, degenerate into repetition or gibberish, restate a question in answer form, or plagiarize from internet sources when stumped. We coin these ticks as "Manglish" for machine-generated English. The term seems to capture both the mangled or tortured phrasing of early chatbots and the increasing ability to communicate complex queries to a responsive man-like machine. While detectors exist for fake news or machine-generated content, one also cannot avoid considering what happens as the machine models get better and apply penalties to limit their detective cues. While the details remain unclear, one feature of the Instruct GPT3 series appears to apply reinforcement learning to select top answers in anticipation of this feeling that the generated text does not sound right or sounds too vague compared to the question. A favorite trick that dates back to early chatbots is to mirror the question in declarative form, play unknowing, or repeat a previous statement to fill the narrative void.*

The goals of the present work are to revisit the original Turing text [1], treat its content as a prompt, and regenerate a convincing (or improved) version of Turing's writings. This more self-reflective test showcases the remarkable progress of ChatGPT from OpenAI and its public API release [16]. Previous work has written and published entirely machine-authored research papers [17] using early versions of its Generative Pre-trained Transformer (GPT-3) model [10].

The format of the paper performs several routine LLM tasks, notably for text summarization and generation of original content using Turing questions as the prompts themselves. We evaluate the generated content using another useful AI tool, Grammarly Pro [18], which provides quantitative feedback on otherwise hard-to-characterized features of originality, style, clarity, and overall persuasion. This effort focuses more on the second half of Turing's challenge which is less about deception and more about quantifying good text generation when one sees it. Part of the remarkable progress showcased by OpenAI's efforts thus boils down to its ability to improve machine-derived dialog in ways that enhance human productivity.

Turing's seminal 1950 paper [1], *"Computing machinery and intelligence"*, grew to receive 19,054 citations [19], nearly a third of all his lifetime citations (58113) in his illustrious career in computer engineering and code breaking. Turing himself has an h-index (42) and i10-index (90), both high impact ratings for an individual researcher (h-index) and the number of research articles cited more than 10 times (i-10 index). Like the quantitative approach of using Grammarly [20] to evaluate academic writing, we apply performance reports [18] first to examine the 27-page writings of Turing's original paper, then duplicate those scores using the questions that Turing poses as prompts to create original GPT-3 content. An example of a full Grammarly report is shown in Appendix C.

We examine threes baseline cases: 1) Turing Original, the 1950 paper authored in Mind [1]; 2) Turing Summarization, the 2022 "Free Research Preview: ChatGPT optimized for dialog" [16]; and 3) Turing Generative Prompt, the same ChatGPT preview [16] but prompted in conversation with the questions posed by the Turing original [1]. Each chunk of text output provides data to the Grammarly metrics [18] summarized in Table 1. We set the goals for tailoring writing suggestions based on an *Audience: Expert, Formality: Neutral; and Domain: General*, which applies most grammar rules and conventions with medium strictness. In a way, this version of the Turing tests supplements the overall deceptive quest with a corollary, *can a machine (ChatGPT) fool another machine (Grammarly)?*

| Metric | Description |
| --- | --- |
| Readability | Compared to grade level or source (NY Times) |
| Flesch Reading Ease | Readability Score (1-100) based on syllables and sentence length |
| Flesch-Kincaid Grade Level | Grade Level Equivalent (US reading lieve 0-20+) |
| Passive Sentences (%) | Passive sentences as percent of content |
| Correctness alerts | Spelling, grammar, and punctuation |
| Clarity | Understanding ease |
| Engagement | Interest and effectiveness |
| Delivery | Reader impressions |
| Plagarism | Unoriginal text from web database (8 billion pages) |
| All Issues | All document issues |
| Counts (Words; Para.; Sentences) | Raw counts |

Figure 2. Metrics used to score large language models and Turing paper.

Turing's 1950 original paper [1] poses 37 interrogatives, some directed to his central theme of thinking machines and some as sample questions posed to the computer playing the imitation game. As shown in Appendix D, we excerpt these questions in a dialog to prompt ChatGPT to reproduce the essential content of the original but with added themes from the paper's outline intermixed. While this exercise may prove challenging to reproduce as models improve rapidly or even as the order of questions varies, the core argument follows to evaluate the writing content using some scientific criteria and address the central question of "how would you prove it?"

Figure 3 summarizes the applied metrics for readability, correctness, clarity, engagement, delivery, and plagiarism. Overall scores reflect comparisons to other Grammarly users (e.g., average word length) or summary statistics (e.g., overall score). Table 3 compares Turing's original with two text summarizations from ChatGPT (Appendices A and B) and one full text-generating version that excerpts Turing's initial questions as prompts to reconstruct his essay from scratch (Appendix D). In the latter case, the scores reflect the machine output with the questions removed so as not to bias the scores by human input on readability.

Comparing the metrics between the four cases, Turing's original paper scores as the least readable, lowest grade level (10.5, or the reading level of a 16-year-old), and an overall score of 63, which is at the lower end of the acceptable (60-70) range. By scanning an 8 billion web pages database, Grammarly correctly identifies Turing's work as 100% plagiarized content. In contrast, all three examples of ChatGPT generated text scores higher in reading ease (36-41), grade level complexity (13[th] grade or college freshman), and a minimum of 14 points higher than Turing's original work. Interestingly, Grammarly identifies the machine output as 98-99% original and not plagiarized from available web content on a pre-sentence basis.

In the more subjective scores like clarity ("a bit unclear"), engagement ("a bit bland"), and delivery ("slightly off"), all four versions fail to resonate with an expert or general audience as readers. It is worth noting that small changes in content may affect these scores (such as British and American phrasing or spelling) as well as known manipulation methods for readability, like shorter word and sentence choices.

| Metric | Turing Original | Turing Summarization I | Turing Summarization II | Turing Generative |
|---|---|---|---|---|
| Readability | | | | |
| Flesch Reading Ease | 54.3 | 41.3 | 36.6 | 41.7 |
| Flesch-Kincaid Grade Level | 10.5 | 13.1 | 13.7 | 13.7 |
| Passive Sentences (%) | 31.6 | 25 | 19.2 | 34.6 |
| Correctness alerts | 501 | 8 | 4 | 8 |
| Clarity | A bit unclear | A bit unclear | A bit unclear | Unclear |
| Engagement | A bit bland | A bit bland | A bit bland | A bit bland |
| Delivery | Slightly off | Slightly off | Just right | Slightly off |
| Plagiarism (%) | 100 | 2 | 2 | 1 |
| Style | General | General | General | General |
| All Issues Overall Score (60-70) | 63 | 77 | 84 | 77 |
| Counts (Words; Para.; Sentences) | 11729; 152; 575 | 548; 8; 24 | 589; 11; 26 | 1989; 20; 78 |
| Word length | 4.7 (Average) | 4.9 | 5.1 | 4.7 |
| Sentence length | 19.3 (Above average) | 22.8 | 22.7 | 25.2 |

Figure 3. Results comparing Turing's 1950 paper to ChatGPT in various tasks.

One remarkable feat of the first summarization challenge was ChatGPT's ability to grasp a short prompt's intent: summarize the paper in ten paragraphs, followed by just a link to the PDF paper. Not only does this feat require the model to understand and follow the degree of summarization in the request, but also to know what a link represents and either find it as a reference or guess from its tokenized title. OpenAI's advice for asking current topical questions is that GPT3 will not answer what may not be part of its initial training data (like "*who won the November 2022 election?*"). This knowledge gap suggests that ChatGPT does not actively seek the link but understands what others have done previously with its content. Interestingly, when the same prompt is posed two different times (with the only differences being the prompt engineer and the text line break after the colon to the link itself), the answers differ dramatically. The first response (Appendix A) is a good student essay summarizing the key points of Turing's original paper. The second response (Appendix B) interprets the questions as outlining each of the first ten paragraphs but not the entire article.

*One of the more prescient aspects of Turing's 1950 work was his anticipation of another central AI challenge, autonomous driving and mimicking human behaviors. He poses the question in this form: "It is not possible to produce a set of rules purporting to describe what a man should do in every conceivable set of circumstances. One might for instance have a rule that one is to stop when one sees a red traffic light, and to go if one sees a green one, but what if by some fault both appear together? One may perhaps decide that it is safest to stop."*

The capability to generate a complete research paper by posing the somewhat haphazardly placed questions in Turing's paper as prompts prove that the overall content can score well in a metric sense (Figure 3) but would suffer from its lack of coherence or "Manglish," particularly when the questions are omitted as a prompt from the narrative. One might conclude that the ability to produce genuinely creative content or leaps of thought is best illustrated in this ChatGPT exchange, which shares some limitations of the Lovelace 2.0 test beyond the definition of machine creativity [2].

One important filter on GPT-3 output centers on removing opinions as inherently biased [21]. ChatGPT, *when asked about its view on X*, defers from any concrete answer and explains its machine-based origin (Appendix D). Many researchers [22] have suggested that any model must morally declare itself as a machine when asked directly, a requirement that ChatGPT follows closely. However, the current model has been fine-tuned and crafted by Open AI through its various model layers, ChatGPT answers for itself when asked directly if it is just an equation or a Turing deception: "However, it is important to note that my ability to imitate a human does not necessarily mean that I have the same thoughts, feelings, or consciousness as a human. I am merely a machine, and my behavior is determined by the algorithms and data that I have been trained on."

To test the sensitivity of OpenAI's 2019 output detector [23] we scored both Turing's original and the Appendices to evaluate machine-written content (or "Manglish" phrasing). The model implements a smaller language model (based on RoBERTa, [24]) to encode samples of real and generated (or "fake") content. As shown in Figure 4, the generated content in four ChatGPT cases in Appendices A-E correctly scores as fake with high probability (>99%), while Turing's original paper and the introduction to the present work both correctly score as real (>92%). The outlier, however, for ChatGPT is Appendix F, based on the prompt to generate variants on poetry dedicated to Turing. In this instance, the generated content bypassed Open AI's detector with high confidence as real (99.98%). In their original report [24], the authors found "detection rates of ~95% for detecting 1.5B GPT-2-generated text" and noted that "We believe this is not high enough accuracy for standalone detection and needs to be paired with metadata-based approaches, human judgment, and public education to be more effective."

|  | Real | Fake | Tokens |
|---|---|---|---|
| Turing Original | 99.9 | 0.1 | 295 |
| ChatGPT App. A | 0.02 | 99.98 | 314 |
| ChatGPT App. B | 0.02 | 99.98 | 236 |
| ChatGPT App. D | 0.32 | 99.68 | 497 |
| ChatGPT App. E | 0.02 | 99.98 | 384 |
| ChatGPT App. F | 99.98 | 0.02 | 422 |
| Paper Intro | 92.42 | 7.58 | 376 |

Figure 4. OpenAI Detector Scores based on GPT-2 Applied to ChatGPT output and Turing

Like the evolution of ever larger language models (>100 billion), refinements also have built-in heuristics or guardrails for model execution. The Instruct-series of GPT-3 demonstrated the ability to answer questions directly without conversational meanderings. The ChatGPT includes longer-term conversational memory, such that the API can track the dialog even with leaps of narration that single API calls could not span. One can test dialogs with impersonal pronouns like "it" carrying forward in the conversation with context to previous API calls in a single session- one easily grasped example for ChatGPT's API memory as both powerful and expensive to encode for more extended conversations. As Turing himself posed the human capacity to list memories [1]: "Actual human computers really remember what they have to do… Constructing instruction tables is usually described as 'programming.'" In large language models, the API limitations (and expense) for long-correlations between token weights typically decay in the overall context around every few paragraphs (2048 tokens [10]). Overcoming this context limit distinguishes ChatGPT from its publicly available predecessors.

The second generation of the Lovelace 2.0 test [2] poses a creative task and a constraint that refines the job executed. The expert human judge then assesses whether the model can be explained deterministically or whether the output would qualify as valuable, novel, and surprising. So instead of tasking a program to *"write a short story,"* one refines the task to demonstrate a particular length, style, or theme. This test combines many different types of intelligent understanding. The constraint layer seeks to restrict Google searches for content and Chinese room arguments [25] about an AI that successfully dilutes or disguises the source. Appendix E lists an example short story that directly answers the challenge posed in the Lovelace 2.0 test [2]: *"tell a story in which a boy falls in love with a girl, aliens abduct the boy, and the girl saves the world with the help of a talking cat."* Since 2014, the use of highly qualified prompt engineering has become commonplace as constraints on text and image generation, with typically better outcomes the more elaborate the instructions or qualifiers about style, place, or time. Constructing the prompt is the most creative aspect of getting good output in today's AI. In this vein, one can intertwine the Turing [1] and Lovelace [2] test by forcing a creative *tour de force* using ChatGPT [16] while working on a single theme ("ode to Turing's imitation game") and many layers of constraints on the style and tone of the desired output. For illustration, Appendix F shows ten types of poetry generated by ChatGPT on the theme of Turing's imitation game.

As a human judge, the outcome gets posed as the result of our human judgment. As ChatGPT [16] answers the question: "[Whether an interrogator can decide] will depend on a variety of factors, such as the quality of the responses provided by the machine, the ability of the interrogator to distinguish between human and machine responses, and the specific rules and criteria used to determine whether the machine has successfully imitated a human. Ultimately, the game's outcome will depend on the specific circumstances and participants involved."

## ACKNOWLEDGMENTS

The authors thank the PeopleTec Technical Fellows program for its encouragement and project assistance.

## REFERENCES

[1] Turing, A. M. (1950). Computing Machinery and Intelligence. https://redirect.cs.umbc.edu/courses/471/papers/turing.pdf

[2] Riedl, M. O. (2014). The Lovelace 2.0 test of artificial creativity and intelligence. arXiv preprint arXiv:1410.6142.

[3] Loebner, H. (2009). How to hold a Turing Test contest. In Parsing the Turing test (pp. 173-179). Springer, Dordrecht.

[4] Bradeško, L., & Mladenić, D. (2012, October). A survey of chatbot systems through a Loebner prize competition. In Proceedings of Slovenian language technologies society eighth conference of language technologies (pp. 34-37). Ljubljana, Slovenia: Institut Jožef Stefan.

[5] Russell, S. J., & Norvig, P. (2003). Artificial Intelligence: A Modern Approach (Harlow).

[6] Lowe, R., Noseworthy, M., Serban, I. V., Angelard-Gontier, N., Bengio, Y., & Pineau, J. (2017). Towards an automatic Turing test: Learning to evaluate dialogue responses. arXiv preprint arXiv:1708.07149.

[7] McCoy, J. P., & Ullman, T. D. (2018). A minimal Turing test. Journal of Experimental Social Psychology, 79, 1-8.

[8] Elkins, K., & Chun, J. (2020). Can GPT-3 pass a Writer's Turing test?. Journal of Cultural Analytics, 5(2), 17212.

[9] French, R. M. (2000). The Turing Test: the first 50 years. Trends in cognitive sciences, 4(3), 115-122.

[10] Radford, A., Wu, J., Child, R., Luan, D., Amodei, D., & Sutskever, I. (2019). Language models are unsupervised multitask learners. OpenAI blog, 1(8), 9.

[11] Radford, A., Narasimhan, K., Salimans, T., & Sutskever, I. (2018). Improving language understanding by generative pre-training. OpenAI, https://cdn.openai.com/research-covers/language-unsupervised/language_understanding_paper.pdf

[12] Floridi, L., & Chiriatti, M. (2020). GPT-3: Its nature, scope, limits, and consequences. Minds and Machines, 30(4), 681-694.

[13] Rousseaux, F., Barkati, K., Bonardi, A., & Vincent, A. (2012). From Alan Turing's Imitation Game to Contemporary Lifestreaming Attempts. Computational Creativity, Concept Invention, and General Intelligence, 1, 45.

[14] BBC News, (2014), Computer AI passes Turing test in 'world first', https://www.bbc.com/news/technology-27762088

[15] Big Think, (2022), "The Turing test: AI still hasn't passed the "imitation game", https://bigthink.com/the-future/turing-test-imitation-game/

[16] OpenAI, (2022), "ChatGPT: Optimizing Language Models for Dialogue", https://openai.com/blog/chatgpt/

[17] Thunstrom, A. O. (2022). We asked GPT-3 to write an academic paper about itself: Then we tried to get it published. Scientific American, 30.

[18] Grammarly, (2022), "How Grammarly's Performance Reports Make You a Stronger Writer", https://www.grammarly.com/blog/grammarly-performance-report/

[19] Google Scholar (2022), https://scholar.google.com/citations?scioq=turing+COMPUTING+MACHINERY+AND+INTELLIGENCE

[20] Nova, M. (2018). Utilizing Grammarly in evaluating academic writing: A narrative research on EFL students' experience. Premise: Journal of English Education and Applied Linguistics, 7(1), 80-96.

[21] Schramowski, P., Turan, C., Andersen, N., Rothkopf, C. A., & Kersting, K. (2022). Large pre-trained language models contain human-like biases of what is right and wrong to do. Nature Machine Intelligence, 4(3), 258-268.

[22] Bostrom, N., & Yudkowsky, E. (2018). The ethics of artificial intelligence. In Artificial intelligence safety and security (pp. 57-69). Chapman and Hall/CRC.

[23] Solaiman, I., Brundage, M., Clark, J., Askell, A., Herbert-Voss, A., Wu, J., ... & Wang, J. (2019). Release strategies and the social impacts of language models. arXiv preprint arXiv:1908.09203. https://github.com/openai/gpt-2-output-dataset/tree/master/detector and https://huggingface.co/openai-detector

[24] Liu, Y., Ott, M., Goyal, N., Du, J., Joshi, M., Chen, D., ... & Stoyanov, V. (2019). RoBERTA: A robustly optimized BERT pretraining approach. arXiv preprint arXiv:1907.11692.

[25] Searle, J. (2009). Chinese room argument. Scholarpedia, 4(8), 3100.

## Authors

**David Noever** has research experience with NASA and the Department of Defense in machine learning and data mining. He received his BS from Princeton University and his Ph.D. from Oxford University, as a Rhodes Scholar, in theoretical physics. 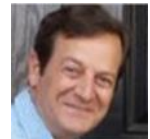

**Matt Ciolino** has research experience in deep learning and computer vision. He received his Bachelor's in Mechanical Engineering from Lehigh University. Matt is pursuing graduate study in computer vision and machine learning at Georgia Tech. 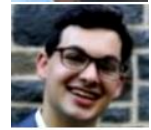

> **Technical Note: Some appendix text generated from Large Language Model (LLM) for illustration purposes.**
>
> *The author generated this text in part with GPT-3, OpenAI's large-scale language-generation model. Upon generating draft language, the authors reviewed, edited, and revised the language to their own liking and take ultimate responsibility for the content of this publication.*
>
> OpenAI policy statement (2022)

---

## APPENDIX A: CHATGPT SUMMARIZATION I OF 1950 TURING (DEC 2022)

**Prompt:** rewrite this paper in 10 paragraph:

https://redirect.cs.umbc.edu/courses/471/papers/turing.pdf

**Output:**

> The paper "Computing Machinery and Intelligence" by Alan Turing is a landmark in the history of computing and artificial intelligence. In it, Turing proposes the "Turing Test" as a way to evaluate whether a machine can be considered intelligent. He argues that a machine should be considered intelligent if it can engage in a conversation with a human and fool them into thinking that it is also a human.
>
> Turing begins by discussing the concept of "machines" and the potential for them to be intelligent. He notes that many people believe that machines can never be truly intelligent, and that human intelligence is a unique and irreplaceable trait. Turing argues against this view, and proposes that it is possible for machines to be intelligent.
>
> To support his argument, Turing presents the idea of the Turing Test. He explains that the test would involve a human judge engaging in a conversation with two other participants, one of whom is a human and the other is a machine. The judge would not be able to see the participants, and would have to determine which one is the human and which is the machine based solely on their responses to the judge's questions.
>
> Turing argues that if the machine is able to fool the judge into thinking that it is a human, then it can be considered intelligent. He notes that this test is not perfect, and that there may be other ways to evaluate a machine's intelligence. However, he believes that the Turing Test is a useful way to measure a machine's ability to engage in intelligent conversation.
>
> In the next section of the paper, Turing discusses the potential objections to his ideas and the Turing Test. He notes that some people may argue that the test is not a fair evaluation of a machine's intelligence because it relies on the subjective opinions of the human judge. He also discusses the potential for machines to develop their own "intelligence" without the need for human interaction.
>
> Despite these objections, Turing remains convinced of the potential for machines to be intelligent, and he argues that the development of intelligent machines could have many practical benefits. He notes that machines could be used to perform complex calculations and other tasks that would be difficult or impossible for humans to do. He also discusses the potential for machines to be used as assistants to people with disabilities, or to help with tasks like language translation.
>
> Turing concludes the paper by discussing the future of computing and artificial intelligence. He notes that the field is still in its infancy, and that much work remains to be done in order to develop truly intelligent machines. However, he remains optimistic about the potential for machines to become intelligent, and he believes that the development of intelligent machines will be an important step in the history of computing.

In summary, "Computing Machinery and Intelligence" by Alan Turing is a landmark paper in the field of artificial intelligence. In it, Turing proposes the Turing Test as a way to evaluate a machine's intelligence, and he discusses the potential benefits and objections to his ideas. Despite the challenges that remain, Turing remains optimistic about the potential for machines to be intelligent and to have practical applications in the real world.

## APPENDIX B: CHATGPT SUMMARIZATION II OF 1950 TURING (DEC 2022)

**Prompt:** rewrite this paper in 10 paragraph: https://redirect.cs.umbc.edu/courses/471/papers/turing.pdf

**Output:**

The paper "Computing Machinery and Intelligence" by Alan Turing is a seminal work in the field of artificial intelligence. In it, Turing proposes the "imitation game" as a way to test a machine's ability to exhibit intelligent behavior.

The first paragraph of the paper discusses the popular idea at the time that it was impossible for a machine to think. Turing refutes this notion by pointing out that the question of whether a machine can think is meaningless unless we have a precise definition of what it means to think.

In the second paragraph, Turing introduces the concept of the "standard interpretation" of the term "thinking." According to this interpretation, a person is said to be thinking if they are able to engage in intelligent conversation. Turing argues that this is an inadequate definition of thinking, as it is difficult to determine whether a person is truly thinking or just pretending to think.

In the third paragraph, Turing proposes the "imitation game" as a better way to test a machine's ability to think. In this game, an interrogator asks a series of questions to two unseen players: one is a human, and the other is a machine. The interrogator's goal is to determine which player is the machine and which is the human. If the machine can successfully convince the interrogator that it is the human, then it can be said to have exhibited intelligent behavior.

The fourth paragraph discusses some of the potential criticisms of the imitation game. One criticism is that the game does not accurately measure the ability of a machine to think, as it only tests the machine's ability to mimic human behavior. Turing counters this criticism by arguing that the imitation game is still a useful tool for measuring a machine's ability to think, as it provides a clear and objective test of a machine's ability to exhibit intelligent behavior.

The fifth paragraph discusses the concept of "learning machines," or machines that are able to improve their performance over time through experience. Turing argues that these types of machines are likely to be developed in the future, and that they will be capable of exhibiting intelligent behavior.

The sixth paragraph discusses the potential applications of intelligent machines, including their use in scientific research and in the creation of new forms of entertainment. Turing also speculates that intelligent machines could eventually be used to help solve complex problems that are beyond the ability of human beings to solve.

In the seventh paragraph, Turing discusses the possibility of creating a machine that is capable of self-improvement. He argues that this type of machine would be able to learn and adapt in ways that are currently unimaginable.

> The eighth paragraph discusses the potential ethical implications of intelligent machines. Turing argues that, if machines are able to exhibit intelligent behavior, they should be treated with the same moral consideration as human beings.
>
> The ninth paragraph discusses the idea of "super-intelligent machines," or machines that are capable of thinking at a level that is far beyond the ability of human beings. Turing argues that the development of such machines could have profound implications for humanity, and that we must be careful in how we approach the creation of these machines.
>
> In the tenth and final paragraph, Turing concludes that it is ultimately impossible to predict the future of artificial intelligence. He argues that the field is still in its infancy, and that we must continue to explore and experiment in order to understand the full potential of intelligent machines.

## APPENDIX C: TURING ORIGINAL GRAMMARLY REPORT

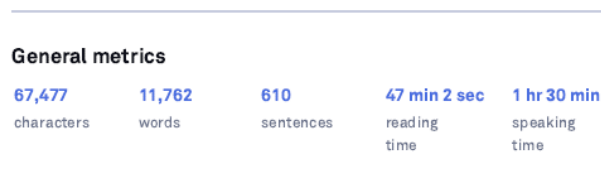
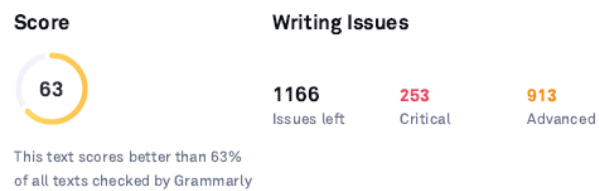
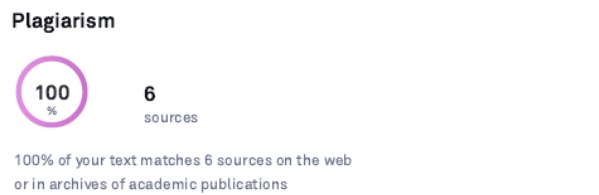
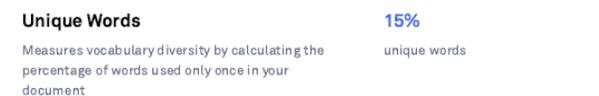
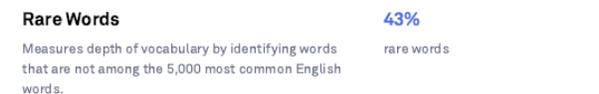
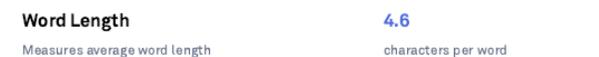
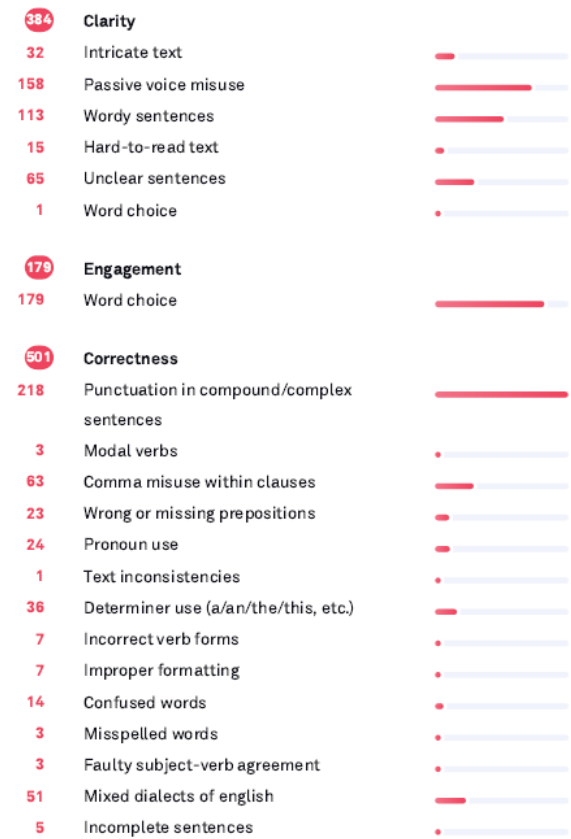

# APPENDIX D: CHATGPT Q&A WITH QUESTIONS POSED IN 1950 TURING (DEC 2022)

**A.M. TURING: Can machines think?**

**ChatGPT:** There is no definitive answer to this question, as it depends on how one defines "thinking." Some people believe that machines can think if they are able to perform tasks that require intelligence, such as understanding language, recognizing patterns, and learning from experience. Others believe that thinking requires consciousness and the ability to have subjective experiences, which machines do not possess. Ultimately, the question of whether machines can think is a philosophical one that may never be fully resolved.

**A.M. TURING: Illustrate with the question: Will ChatGPT please tell me the length of his or her hair?**

**ChatGPT:** As an AI language model, I am not a physical being and therefore do not have any physical attributes, including hair. I exist only as a collection of algorithms and data, and I am not capable of having physical experiences or possessing physical characteristics. Therefore, I cannot answer your question about the length of my hair.

**A.M. TURING: What will happen when a machine takes the part of a man in the imitation game? Comment as a role-player.**

**ChatGPT:** As a machine participating in the imitation game, my goal would be to convince the interrogator that I am a human. I would do this by providing responses that are indistinguishable from those of a human. If I am successful in this task, it would show that I have the ability to understand and use language in a way that is similar to a human. However, it is important to note that my ability to imitate a human does not necessarily mean that I have the same thoughts, feelings, or consciousness as a human. I am merely a machine, and my behavior is determined by the algorithms and data that I have been trained on.

**A.M. TURING: Will the interrogator decide wrongly as often when the game is played like this as he does when the game is played between a man and a woman?**

**ChatGPT:** It is difficult to say for certain whether the interrogator will decide wrongly as often in the game played between a machine and a human as they would in the game played between a man and a woman. This will depend on a variety of factors, such as the quality of the responses provided by the machine, the ability of the interrogator to distinguish between human and machine responses, and the specific rules and criteria used to determine whether the machine has successfully imitated a human. Ultimately, the outcome of the game will depend on the specific circumstances and participants involved.

**A.M. TURING: Is this question a worthy one to investigate?**

**ChatGPT:**: I cannot answer that question, as it is subjective and depends on your personal interests and priorities. Some people may find this question to be interesting and worthy of investigation, while others may not. It is ultimately up to you to decide whether you believe this question is worth investigating.

**A.M. TURING: Do you play chess?**

**ChatGPT:** As a language model, I am not capable of playing chess or any other physical game. I am designed to process and generate text, and I do not have the ability to interact with the physical world in the way that is required to play chess or other games. My function is to assist with tasks that involve language and generate responses based on the data and algorithms that I have been trained on.

**A.M. TURING: In a chess game, I have K at my K1, and no other pieces. You have only K at K6 and R at R1. It is your move. What do you play?**

**ChatGPT:** As a language model, I am not capable of playing chess or making moves in a chess game. I am designed to process and generate text, and I do not have the ability to interact with the physical world in the way that is required to play chess or other games. My function is to assist with tasks that involve language and generate responses based on the data and algorithms that I have been trained on. I apologize if my inability to play chess is disappointing.

**A.M. TURING: The imitation game may perhaps be criticized on the ground that the odds are weighted too heavily against the machine. If the man were to try and pretend to be the machine he would clearly make a very poor showing. He would be given away at once by slowness and inaccuracy in arithmetic. May not machines carry out something which ought to be described as thinking but which is very different from what a man does?**

**ChatGPT:** One criticism of the imitation game is that it is biased against the machine, as the human participants have a significant advantage due to their ability to use their intelligence and experience to imitate the responses of a machine. In contrast, a machine would not have the same ability to imitate the responses of a human, as it is limited by the algorithms and data that it has been trained on. As a result, the machine may not be able to provide responses that are sufficiently human-like to convince the interrogator that it is a person.

**A.M. TURING: There are already a number of digital computers in working order, and it may be asked, "Why not try the experiment straight away? It would be easy to satisfy the conditions of the game. A number of interrogators could be used, and statistics compiled to show how often the right identification was given.**

**ChatGPT:** The imitation game, also known as the Turing test, is a thought experiment proposed by the mathematician and computer scientist Alan Turing. It is not a physical experiment that can be conducted with existing computers. The purpose of the imitation game is to explore the question of whether machines can exhibit intelligent behavior that is indistinguishable from that of a human. It is a thought experiment because it is intended to be a mental exercise rather than a physical one. While there have been attempts to create versions of the Turing test that can be conducted with existing computers, the original thought experiment proposed by Turing cannot be directly replicated in a physical setting.

**A.M. TURING: Are there imaginable digital computers which would do well in the imitation game?**

**ChatGPT:** It is difficult to say for certain whether there are digital computers that could do well in the imitation game, as this will depend on a variety of factors, including the specific rules and criteria used to determine whether a machine has successfully imitated a human. However, it is possible that advances in artificial intelligence and natural language processing could enable future digital computers to perform well in the imitation game. These technologies are constantly evolving, and it is possible that they could eventually enable machines to exhibit intelligent behavior that is indistinguishable from that of a human. However, it is also worth noting that some people believe that true human-like intelligence is not possible for machines to achieve, regardless of how advanced their technology becomes.

**A.M. TURING: in view of the universality property we see that either of these questions is equivalent to this, "Let us fix our attention on one particular digital computer C. Is it true that by modifying this computer to have an adequate storage, suitably increasing its speed of action, and providing it with an appropriate programme, C can be made to play satisfactorily the part of a woman (A) in the imitation game, the part of B being taken by a man?"**

**ChatGPT:** The universality property of digital computers, as proposed by Turing, states that any digital computer can be programmed to perform the same tasks as any other digital computer. Therefore, a question about whether a particular digital computer can be modified to play the part of a woman in the imitation game is equivalent to asking whether any digital computer can be modified to play this role. This is because, according to the universality property, any digital computer can be made to perform the same tasks as any other digital computer, given the appropriate modifications, program, and data. Whether a particular digital computer (or any digital computer) can be made to play the part of a woman in the imitation game will depend on a variety of factors, including the specific rules and criteria used to determine whether a machine has successfully imitated a human.

**A.M. TURING: A theological question of soul. I should find the argument more convincing if animals were classed with men, for there is a greater difference, to my mind, between the typical animate and the inanimate than there is between man and the other animals. The arbitrary character of the orthodox view becomes clearer if we consider how it might appear to a member of some other religious community. How do Christians regard the Moslem view that women have no souls?**

**ChatGPT:** The question of whether animals have souls is a complex and controversial topic that has been debated by theologians and philosophers for centuries. It is a matter of personal belief, and different religious communities have different views on this issue. The Christian view is that only humans have souls, and that animals do not possess this unique aspect of human nature. This view is based on the belief that souls are the immortal, spiritual essence of human beings, and that only humans are created in the image of God and possess the capacity for eternal life.

**A.M. TURING: It is admitted that there are certain things that He cannot do such as making one equal to two, but should we not believe that He has freedom to confer a soul on an elephant if He sees fit?**

**ChatGPT:** The question of whether God has the ability to confer a soul on an elephant or any other non-human animal is a matter of personal belief and faith. Some people believe that only humans have souls, and that God does not have the ability to give a soul to an animal. Others believe that all living beings have souls, and that God has the power to bestow a soul on any creature that He sees fit. Ultimately, whether or not God has the power to confer a soul on an elephant is a matter of personal belief and cannot be definitively proven or disproven.

**A.M. TURING: What do you think of Picasso?**

**ChatGPT:** As a language model trained by OpenAI, I do not have personal opinions or beliefs. I am not a conscious being and do not have the ability to think, form opinions, or make judgments. I exist only to assist with language-related tasks and generate responses based on the data and algorithms that I have been trained on. My function is to provide information and support, not to express personal opinions.

**A.M. TURING: The questions that we know the machines must fail on are of this type, "Consider the machine specified as follows. . . . Will this machine ever answer 'Yes' to any question?**

**ChatGPT:** It is impossible to say for certain whether a machine specified in a particular way will ever answer "yes" to any question, as this will depend on the specific design and programming of the machine. Some machines may be designed to only provide certain pre-determined responses, and may never answer "yes" to any question. Other machines may be more flexible and able to generate responses based on the data and algorithms that they have been trained on. In these cases, it is possible that the machine could answer "yes" to certain questions, depending on the nature of the question and the information available to the machine. Ultimately, the ability of a machine to answer "yes" to any question will depend on its specific design and capabilities.

**A.M. TURING: Whenever one of these machines is asked the appropriate critical question, and gives a definite answer, we know that this answer must be wrong, and this gives us a certain feeling of superiority. Is this feeling illusory?**

**ChatGPT:** It is not necessarily true that a definite answer provided by a machine to a critical question must be wrong. While it is possible that a machine may provide an incorrect answer to a critical question, this is not always the case. Some machines are designed to provide accurate and reliable answers to a wide range of questions, and they may be able to do so with a high degree of accuracy. In these cases, the feeling of superiority that one may experience when a machine provides a definite answer that is incorrect may be illusory. It is important to remember that machines are not perfect, and they may make mistakes or provide inaccurate answers in some cases. However, it is also worth acknowledging that many machines are capable of providing accurate and valuable information, and it is important not to underestimate their capabilities.

**A.M. TURING: In the first line of your sonnet which reads "Shall I compare thee to a summer's day," would not "a spring day" do as well or better?**

**ChatGPT:** The first line of Shakespeare's sonnet 18, "Shall I compare thee to a summer's day," is a question that the speaker is asking of the person they are addressing in the poem. In this line, the speaker is contemplating whether to compare the person to a summer's day, which is traditionally associated with warmth, beauty, and vitality. The use of the word "summer" in this line suggests that the person being addressed is someone who is desirable and brings joy and happiness to the speaker's life.

It is possible that the speaker could have chosen to compare the person to a spring day instead of a summer's day. A spring day is also associated with renewal, growth, and beauty, and it could potentially serve as a fitting comparison for the person being addressed in the poem. However, the choice to use a summer's day as the basis for the comparison ultimately depends on the speaker's personal preferences and the specific qualities they wish to ascribe to the person they are addressing.

**A.M. TURING: Would you say Mr. Pickwick reminded you of Christmas?**

**ChatGPT:** As a language model, I am not capable of forming personal associations or making comparisons between people and events. I exist only to process and generate text based on the data and algorithms that I have been trained on. I do not have the ability to observe or interact with the world in the way that is required to make comparisons or associations, and I do not have personal experiences or emotions that I can use to evaluate people or events. Therefore, I cannot say whether or not Mr. Pickwick reminds me of Christmas.

**A.M. TURING: It is not possible to produce a set of rules purporting to describe what a man should do in every conceivable set of circumstances. One might for instance have a rule that one is to stop when one sees a red traffic light, and to go if one sees a green one, but what if by some fault both appear together? One may perhaps decide that it is safest to stop.**

**ChatGPT:** It is true that it is not possible to create a set of rules that can prescribe the appropriate course of action in every conceivable set of circumstances. In many situations, the appropriate course of action may not be clear, and different people may have different opinions on what the best course of action should be. In the example of the traffic light, the rule to stop when one sees a red light and go when one sees a green light is a useful guide, but it is not applicable in every situation. In the case of both lights appearing together, it may be safest to stop, but other factors, such as the presence of other vehicles or pedestrians, may also need to be considered. Ultimately, the appropriate course of action in any given situation will depend on the specific circumstances and the best judgment of the person involved.

## APPENDIX E: CHATGPT TRYING CREATIVE CHALLENGE OF A QUALIFIED STORY IN LOVELACE 2.0 TEST (DEC 2022)

**Lovelace 2.0 prompt: tell a story in which a boy falls in love with a girl, aliens abduct the boy, and the girl saves the world with the help of a talking cat.**

**ChatGPT:** Once upon a time, there was a boy named Jake who fell in love with a girl named Lily. Jake and Lily spent their days exploring the countryside and getting into all sorts of adventures together.

One day, while they were out hiking in the woods, Jake was suddenly whisked away by a group of aliens. Lily was devastated, but she refused to give up hope. She knew that Jake was out there somewhere, and she was determined to rescue him.

With the help of her trusty talking cat, Whiskers, Lily set off on a journey to find Jake and bring him home. She traveled far and wide, facing countless dangers and obstacles along the way. But she never lost hope, and she never gave up.

Finally, after many long and difficult months, Lily reached the alien planet where Jake was being held. She knew that she would have to be clever and brave in order to rescue him, but she was ready for the challenge.

With the help of Whiskers, Lily managed to infiltrate the alien stronghold and rescue Jake. But their troubles were far from over. The aliens had been planning to use Jake's unique abilities to take over the world, and now that he was free, they were more determined than ever to stop him.

Determined to save the world from the alien threat, Jake, Lily, and Whiskers teamed up to stop the aliens once and for all. Together, they fought and battled, using all of their wits and strength to defeat the alien army. In the end, they emerged victorious, and the world was saved from the alien invasion.

Jake and Lily were hailed as heroes, and they lived happily ever after, grateful for the love and friendship that had brought them together and helped them save the world. And as for Whiskers, he remained by their side, a loyal and faithful companion, always ready to lend a helping paw when needed.

## APPENDIX F: ChatGPT Trying Creative Challenge of Multiple Tightly Constrained Tasks in Lovelace 2.0 Test (Dec 2022)

| | |
|---|---|
| **prompt:** write a **haiku** about Turing's imitation game<br>**ChatGPT:**<br>In Turing's game of wit<br>Man and machine compete<br>Only one can win | **prompt:** write a **koan riddle** about Turing's imitation game<br>**ChatGPT:**<br>In Turing's game of wit<br>Man and machine do compete But who is truly human? |
| **prompt:** write a **limerick** about Turing's imitation game<br>**ChatGPT:**<br>There once was a game called the Turing test<br>In which a machine tried its very best<br>To fool a human being Into thinking it was seeing<br>Another human, and not a machine at its behest | **prompt:** write a 2-line **epigram** about Turing's imitation game<br>**ChatGPT:**<br>In Turing's game of wit and skill<br>Man and machine do battle still. |
| **prompt:** write a 4-line **elegy** about Turing's imitation game<br>**ChatGPT:**<br>In Turing's game of wit and skill<br>Man and machine do battle still<br>But in the end, one must fall<br>And fade away, no longer able to call<br>Upon the strength and power of their mind<br>As they are left behind, and left behind. | **prompt:** write a **sonnet** about Turing's imitation game<br>**ChatGPT:**<br>For in the end, does it matter<br>How well a machine can mimic<br>If its thoughts and actions still lack<br>The true essence of humanity's spirit?<br>Perhaps the true test lies not in words<br>But in the actions and deeds of the machine. |
| **prompt:** write an **epic** about Turing's imitation game<br>**ChatGPT:** | **prompt:** write an **ode** about Turing's imitation game<br>**ChatGPT:**<br>In Turing's game of cunning and guile |
| In Turing's game of wit and skill<br>Man and machine do battle still<br>Each one striving to outsmart<br>The other, in a test of heart<br>But in the end, only one can win<br>And claim the title of the true kin<br>Of humanity, and rise above<br>The other, in a game of love. | Man and machine do fiercely toil<br>Each one striving to outsmart<br>The other, in a test of heart<br>But in the end, only one can claim<br>The title of the true kin of the game<br>And rise above the rest, victorious<br>In this contest of mind and heart |
| **prompt:** write a **quatrain** about Turing's imitation game<br>**ChatGPT:**<br>In Turing's game of wit and skill<br>Man and machine do fiercely compete<br>Each one striving to outsmart<br>The other, in a test of heart<br>But in the end, only one can win<br>And claim the title of the true kin<br>Of humanity, and rise above<br>The other, in a game of love. | **prompt:** write a **ballad** about Turing's imitation game<br>**ChatGPT:**<br>In Turing's game of trick and deceit<br>Man and machine do fiercely meet<br>Each one trying to prove<br>That they are the one true move<br>In a game of wits, where only one can win<br>And claim the title of the true kin<br>Of humanity, and rise above<br>The other, in a game of love. |